\pdfoutput=1

\documentclass[11pt]{article}

\usepackage[final]{acl}
\PassOptionsToPackage{hyphens}{url}\usepackage{hyperref}
\usepackage{amsmath}
\usepackage{amssymb}
\usepackage{bbm}
\usepackage{times}
\usepackage{latexsym}
\usepackage{stfloats}
\usepackage{multirow}
\usepackage{multicol}
\usepackage{subcaption}
\usepackage{graphicx}
\usepackage{tikz}
\usepackage{environ}
\usepackage{pifont}
\usepackage{xcolor}
\usepackage[normalem]{ulem}
\usepackage{booktabs}
\usepackage[T1]{fontenc}

\usepackage[utf8]{inputenc}

\usepackage{microtype}

\definecolor{GreenBackground}{RGB}{228, 240, 217}
\definecolor{Green}{RGB}{144, 181, 0}
\definecolor{RedBackground}{RGB}{248, 230, 213}
\definecolor{Red}{RGB}{241, 127, 0}
\definecolor{BlueBackground}{RGB}{224, 235, 247}
\definecolor{Blue}{RGB}{82, 115, 197}
\definecolor{GreyBackground}{RGB}{237, 237, 237}
\definecolor{Grey}{RGB}{155, 155, 155}
\definecolor{YellowBackground}{RGB}{253, 245, 215}
\definecolor{Yellow}{RGB}{243, 213, 126}
\definecolor{PurpleBackground}{RGB}{209, 204, 237}
\definecolor{Purple}{RGB}{148, 0, 211}

\title{GLaPE: Gold Label-agnostic Prompt Evaluation for \\ Large Language Models}

\author{Xuanchang Zhang, Zhuosheng Zhang, Hai Zhao \\
        Shanghai Jiao Tong University \\
        \texttt{\{zxcsamzxc, zhangzs\}@sjtu.edu.cn, zhaohai@cs.sjtu.edu.cn}}

\author{Xuanchang Zhang$^{1}$, Zhuosheng Zhang$^{2,*}$, Hai Zhao$^{3,4,5,}$\thanks{Corresponding authors. This research was supported by the Joint Research Project of Yangtze River Delta Science and Technology Innovation Community (No. 2022CSJGG1400) and National Natural Science Foundation of China (No. 62406188).}
\\
$^1$College of Zhiyuan, Shanghai Jiao Tong University\\$^2$School of Electronic Information and Electrical Engineering, Shanghai Jiao Tong University\\ $^3$Department of Computer Science and Engineering, Shanghai Jiao Tong University
  \\ $^4$Key Laboratory of Shanghai Education Commission for Intelligent Interaction \\
and Cognitive Engineering, Shanghai Jiao Tong University \\ $^5$Shanghai Key Laboratory of Trusted Data Circulation and Governance in Web3 \\ 
\texttt{\{zxcsamzxc, zhangzs\}@sjtu.edu.cn, zhaohai@cs.sjtu.edu.cn}
}

\begin{document}
\maketitle

\begin{abstract}
Despite the rapid progress of large language models (LLMs), their task performance remains sensitive to prompt design.
Recent studies have explored leveraging the LLM itself as an optimizer to identify optimal prompts that maximize task accuracy.
However, when evaluating prompts, such approaches heavily rely on elusive manually annotated gold labels to calculate task accuracy for each candidate prompt, which hinders its generality.
To overcome the limitation, this work proposes GLaPE, a gold label-agnostic prompt evaluation method to alleviate dependence on gold labels.
GLaPE is composed of two critical aspects: self-consistency evaluation of a single prompt and mutual-consistency refinement across multiple prompts. 
Experimental results on 8 widely-recognized reasoning tasks demonstrate that GLaPE can produce more effective prompts, achieving performance comparable to those derived from manually annotated gold labels.
Analysis shows that GLaPE provides reliable evaluations aligned with accuracy, even in the absence of gold labels.
Code is publicly available at \url{https://github.com/thunderous77/GLaPE}.


\end{abstract}

\section{Introduction}

As the integration of large language models (LLMs) into natural language processing tasks has become imperative in recent years \citep{openai2023gpt4, workshop2023bloom, chowdhery2022palm, touvron2023llama}, the sensitivity of the performance of LLMs to prompts has garnered significant attention \citep{pezeshkpour2023large, Loya_2023}. 
While traditional soft prompt tuning methods \citep{li2021prefix, liu2022p, lester2021power, qin2021learning} demonstrate effectiveness in guiding the LLM to perform desired tasks, they encounter limitations when applied to private LLMs, such as GPT-4 \citep{gpt4}. 
This situation necessitates the exploration of effective strategies for optimizing prompts without requiring gradient updates. 

\begin{figure*}[t]
    \centering
    
    \begin{tikzpicture}
        \tikzstyle{mybox} = [draw=white, fill=GreyBackground, thin,
            rectangle, rounded corners, inner sep=5pt, inner ysep=5pt]
        
        \node [mybox] (box1) at (-5.0, 2.2){
            \begin{minipage}{0.44\textwidth}
                \centering
                \textbf{\small (a) Accuracy Evaluation}
                
                \vspace{5pt}
                \tiny
                \raggedright

                \textbf{Question:} Oscar has 24 lollipops and eats 2 on his way to school. He passes 14 out to his friends. He buys twice as many lollipops on his way home as he gave to his friends. He eats 3 more that night and 2 more in the morning.  How many lollipops does Oscar have?\\
                \textcolor{Blue}{\textbf{Gold Label (Answer):} 31}
                
                \textbf{Prompt1:} By carefully analyzing all aspects of the situation, the optimal solution becomes crystal clear.\\
                \textbf{Responses:} \textcolor{Blue}{31, 31, 31, 31, 31, 31, 31, 31, 31, 31} \textcolor{Green}{\ding{52}}\\
                \textbf{Score:} \textcolor{Green}{100.0}\\
                
                \textbf{Prompt2:} After thorough examination and careful consideration, the optimal solution becomes clear.\\
                \textbf{Responses:} 19, \textcolor{Blue}{31, 31, 31, 31, 31, 31, 31}, 36, 36 \textcolor{Green}{\ding{52}}\\
                \textbf{Score:} \textcolor{Green}{100.0}\\
                
                \textbf{Prompt3:} Let's think about this logically.\\
                \textbf{Responses:} \textcolor{Blue}{31}, 33, 33, 36, 36, 36, 36, 36, 36, 36 \textcolor{red}{\ding{56}}\\
                \textbf{Score:} \textcolor{red}{0.0}\\
                
                \textbf{Prompt4:} Let's approach this problem systematically.\\
                \textbf{Responses:} 19, \textcolor{Blue}{31, 31, 31}, 33, 33, 36, 36, 36, 36 \textcolor{red}{\ding{56}}\\
                \textbf{Score:} \textcolor{red}{0.0}\\
                
                \textbf{Prompt5:} By carefully analyzing all the available data, the optimal solution becomes unequivocally evident.\\
                \textbf{Responses:} 8, 8, 11, 19, 19, \textcolor{Blue}{31, 31}, 36, 36, 36 \textcolor{red}{\ding{56}}\\
                \textbf{Score:} \textcolor{red}{0.0}\\
            \end{minipage}
        };

        \node [mybox] (box2) at (2.5,2.2){ 
                \begin{minipage}{0.44\textwidth}
                \centering
                \textbf{\small (b) Our GLaPE Method}
                 
                \vspace{5pt}
                \tiny
                \raggedright

                \textbf{Question:} Oscar has 24 lollipops and eats 2 on his way to school. He passes 14 out to his friends. He buys twice as many lollipops on his way home as he gave to his friends. He eats 3 more that night and 2 more in the morning.  How many lollipops does Oscar have?\\
                \sout{\textcolor{Blue}{\textbf{Gold Label (Answer):} 31}}
                
                \textbf{Prompt1:} By carefully analyzing all aspects of the situation, the optimal solution becomes crystal clear. \\
                \textbf{Responses:} \textcolor{Yellow}{31, 31, 31, 31, 31, 31, 31, 31, 31, 31} \\
                \textbf{Score:} \textcolor{Green}{87.9}\\
                
                \textbf{Prompt2:} After thorough examination and careful consideration, the optimal solution becomes clear.\\
                \textbf{Responses:} 19, \textcolor{Yellow}{31, 31, 31, 31, 31, 31, 31}, 36, 36 \\
                \textbf{Score:} \textcolor{Green}{81.8}\\
                
                \textbf{Prompt3:} Let's think about this logically.\\
                \textbf{Responses:} 31, 33, 33, \textcolor{Yellow}{36, 36, 36, 36, 36, 36, 36} \\
                \textbf{Score:} \textcolor{red}{50.0}\\
                
                \textbf{Prompt4:} Let's approach this problem systematically.\\
                \textbf{Responses:} 19, 31, 31, 31, 33, 33, \textcolor{Yellow}{36, 36, 36, 36} \\
                \textbf{Score:} \textcolor{red}{45.7}\\
                
                \textbf{Prompt5:} By carefully analyzing all the available data, the optimal solution becomes unequivocally evident.\\
                \textbf{Responses:} 8, 8, 11, 19, 19, 31, 31, \textcolor{Yellow}{36, 36, 36} \\
                \textbf{Score:} \textcolor{red}{44.2}\\
            \end{minipage}
        };

        \node [mybox] (box3) at (-1.3, 7.0){
            \begin{minipage}{0.91\textwidth}
                \centering
                \textbf{\small Meta-prompt for Prompt Optimization}
                 
                \vspace{5pt}
                \tiny
                \raggedright

                \textbf{Meta-prompt 1} like ``I have some texts along with their corresponding scores.''\\
                \vspace{5pt}

                Prompt: ... Score: ... ; Prompt: ... Score: ... ; [\textbf{more prompts and scores}]\\
                \vspace{7pt}
                \textbf{Meta-prompt 2} like ``Write your new text that is different from the old ones and has a score as high as possible.''
            \end{minipage}
        };
        \node [mybox,draw=PurpleBackground,text width=220.0,fill=PurpleBackground] (box4) at (-4.6,6.74)  {\tiny Prompt: ... Score: ... ; Prompt: ... Score: ... ; [\textbf{more prompts and scores}]};

        \node [mybox,draw=white, fill=white] (box4) at (-3.5, 5.72)  {\small\textcolor{Purple}{Prompt Evaluation}};
        \node [mybox,draw=white, fill=white] (box4) at (4.0, 5.72)  {\small\textcolor{Purple}{Prompt Evaluation}};

        \draw[-stealth,very thick, Purple](-5,5.5)--(-5, 6.0) ;
        \draw[-stealth, very thick, Purple](2.5, 5.5)--(2.5, 6.0);

        \draw[-stealth, very thick, rounded corners](-5.0, 8.0)--(-5.0,8.5)--(-9.2,8.5)--(-9.2,2.5)--(-8.8,2.5);

        \draw[-stealth, very thick, rounded corners](3.0, 8.0)--(3.0,8.5)--(6.7,8.5)--(6.7,2.5)--(6.3,2.5);

        \node [mybox,draw=white, fill=white] (box4) at (-7.2, 8.9)  {\small New Prompts};
        \node [mybox,draw=white, fill=white] (box4) at (5.2, 8.9)  {\small New Prompts};
        
    \end{tikzpicture}

    \vspace{2mm}

    \caption{Sketch of prompt optimization utilizing the LLM as an optimizer \citep{LLMasOPT}, featuring distinct prompt evaluation metrics based on: (a) accuracy or (b) our proposed GLaPE. The texts are favorably read in colors. \textcolor{Blue}{Blue}: gold label, \textcolor{Yellow}{Yellow}: most frequent answer, \textcolor{Green}{Green}: high score, \textcolor{red}{Red}: low score, \textcolor{Purple}{Purple}: prompt evaluation.}
    \label{fig:sketch of LLMasOPT}
    \vspace{0mm}
\end{figure*}
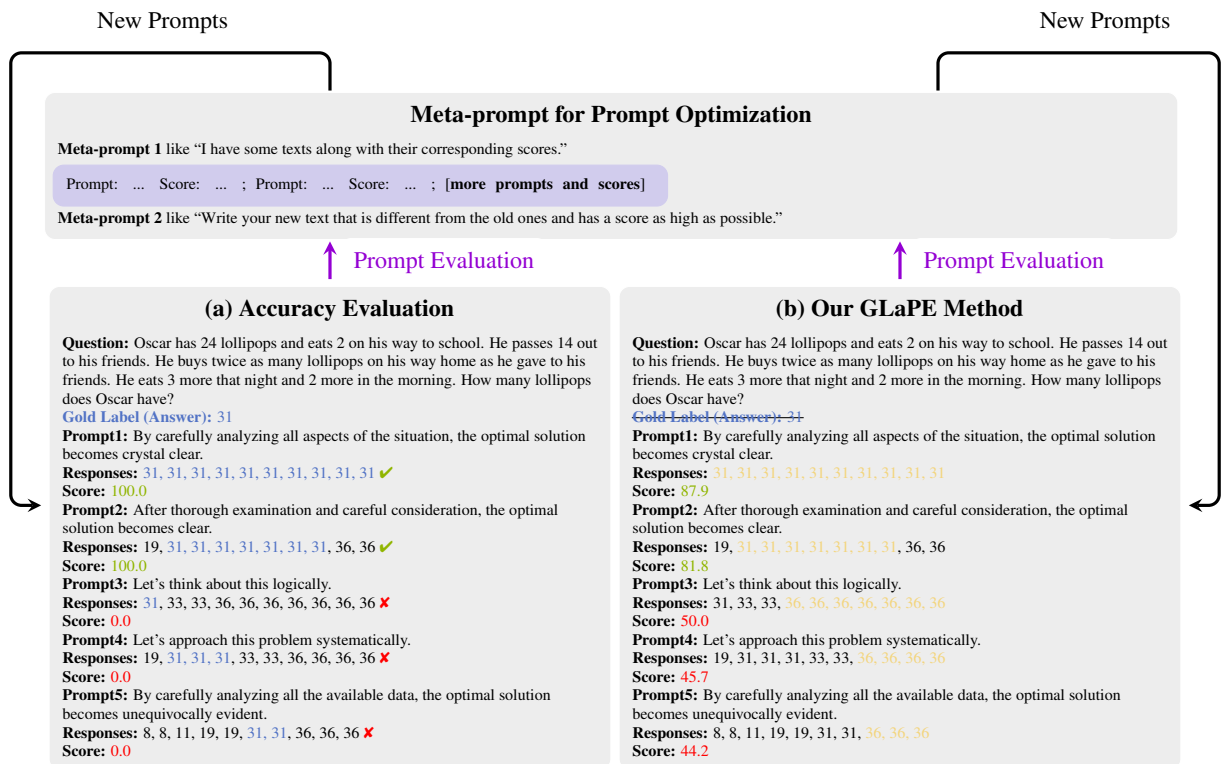

Recent studies \citep{LLMasOPT, zhou2022large} have unveiled a noteworthy strategy, where the LLM itself acts as the optimizer to seek the prompt that maximizes task accuracy. 
Specifically, OPRO \citep{LLMasOPT} provides an intriguing avenue for prompt optimization based on a gold label evaluation recipe (Figure~\ref{fig:sketch of LLMasOPT}a). The optimization commences with an initial prompt, then iteratively evaluates existing prompts and generates novel prompts based on prior assessments. 
However, a significant caveat emerges as these studies heavily rely on manually annotated gold labels.
Concretely, the gold label, representing the ideal output, serves as a crucial ingredient for evaluating and refining prompts. 
Nevertheless, the acquisition of such gold labels poses a formidable obstacle \citep{huang2023large, stechly2023gpt}, introducing complexity and hindering the widespread implementation and generality of these optimization techniques. 
Therefore, exploring alternative methodologies becomes mandatory to address these challenges and improve the efficiency of prompt evaluation and optimization for LLMs.

To address the limitations, this work proposes a gold label-agnostic prompt evaluation (GLaPE) method to identify prompts that facilitate consistent and accurate answers. 
Instead of relying on gold labels, GLaPE evaluates prompts based on two critical aspects: self-consistency evaluation and mutual-consistency refinement. 
Inspired by \citet{wang2022self}, we first consider a naive solution by utilizing self-consistency (SC) as the evaluation metric instead of accuracy, as correct answers generally exhibit higher SC than incorrect ones. 
However, we will show that SC alone may not always yield accurate evaluations, since SC does not always align well with accuracy and can overestimate prompts that produce incorrect but consistent answers. 
To mitigate this, we then propose a complementary approach named mutual-consistency refinement across multiple prompts. 
This approach penalizes inconsistent scores based on SC across prompts that produce the same answers. By doing so, the refinement process effectively identifies prompts that demonstrate high SC but result in incorrect answers, leading to more reliable evaluation scores.
Figure~\ref{fig:sketch for our GLaPE} illustrates our GLaPE method.

Building on our GLaPE evaluation strategy, we then develop a gold label-agnostic prompt optimization method. Specifically, we substitute the accuracy evaluation method in OPRO with our GLaPE method (Figure~\ref{fig:sketch of LLMasOPT}b). Experimental results on 8 widely-recognized reasoning tasks demonstrate that GLaPE can produce more effective prompts, achieving performance comparable to those derived from manually annotated gold labels.

Our key contributions are as follows:

(i) This work studies a gold label-agnostic prompt evaluation method to alleviate dependence on gold labels, which allows prompt evaluation in more realistic scenarios when human-annotated dataset is unavailable. To the best of our knowledge, this work is the first to study gold label-agnostic prompt evaluation for LLMs. 

(ii) We propose a novel prompt evaluation approach named GLaPE, which consists of self-consistency evaluation of a single prompt and mutual-consistency refinement across multiple prompts. GLaPE helps LLMs optimize effective prompts that are comparable with those derived from manually annotated gold labels.

(iii) We elicit the analysis of why the widely-used SC approach fails at our evaluation task and figure out an effective mutual-consistency refinement approach to mitigate the challenge. 




\section{Related Work}

\textbf{Prompt Optimization} In the domain of LLMs \citep{openai2023gpt4, workshop2023bloom, chowdhery2022palm, touvron2023llama}, prompt engineering plays a crucial role in guiding models to generate desired outputs across diverse tasks \citep{pezeshkpour2023large, Loya_2023}. Consequently, optimizing prompts becomes paramount for enhancing the performance and efficiency of LLMs. Various soft prompt tuning methods \citep{li2021prefix, liu2022p, lester2021power, qin2021learning} have been explored in previous research to optimize prompts for open-source LLMs. However, these methods encounter challenges when applied to private LLMs, where accessing gradients is infeasible. Consequently, diverse gradient-free prompt optimization techniques \citep{zhou2022large,pan2023plum,ye2023prompt} have been explored. Recent works \citep{LLMasOPT} have embraced an iterative process for gradient-free prompt optimization, commencing from an initial prompt and iteratively assessing existing prompts while generating new ones based on prior evaluations. Nevertheless, these iterative prompt optimization methods heavily depend on gold labels for prompt evaluation. 
Relying on gold labels not only limits the generalizability of these prompt optimization methods but may also introduce other potential issues~\citep{huang2023large,stechly2023gpt}.
In our work, we propose a novel gold label-agnostic prompt evaluation method and subsequently present a unique approach to optimize prompts for LLMs without the constraints associated with conventional gold label reliance.

\textbf{Prompt Selection} Prompt selection tasks aim to identify the optimal prompt among candidates for a given task, representing an alternative approach to prompt optimization. Recent studies have delved into probability-based evaluation methods, utilizing diverse metrics such as mutual information \citep{Sorensen_2022}, entropy \citep{lu2021fantastically}, and perplexity \citep{gonen2022demystifying}. In contrast to these probability-centric assessments, our proposed evaluation approach exclusively relies on the output, making it applicable to private LLMs where only the output is accessible.

\section{Background}

\subsection{Task Formulation}\label{sec:task_def}
Existing studies on prompt design \citep{LLMasOPT, zhou2022large} generally adhere to a two-stage paradigm in an iterative manner: (i) evaluate the prompt, analogous to calculating the loss function and gradient in soft prompt tuning; (ii) optimize the prompt, analogous to the gradient descent process in soft prompt tuning.

We formulate the two stages on top of the widely-used question-answering (QA) task defined by QA pairs $(Q, A)$, where each pair comprises an input $Q$ and its corresponding expected output $A$. We introduce the prompted model as $\mathcal{M}$ and an evaluation function $f$. Our objective is to determine the optimal natural language instruction prompt. 




To begin with, we define the \textit{meta-prompt} as the input to for prompt optimization. As the upper block shown in Figure \ref{fig:sketch of LLMasOPT}, a meta-prompt contains three parts. The first part is a problem description. The second part is an optimization trajectory, includes past solutions and their evaluation scores. The third part is the optimization instruction for generating new candidate prompts.

Then, we describe the process of obtaining the optimization trajectory. In each iteration, the LLM generates a candidate prompt $\rho$ to the QA task. We concatenate each question $Q$ with the candidate prompt $\rho$ to form the prompted input $[Q; \rho]$. Then, the prompted input is feed to the model to obtain the response $\mathcal{M}\left([Q; \rho]\right)$. We evaluate the goodness of candidate prompt $\rho$ based on the evaluation function $f$, e.g., calculating the accuracy between each pair of $\mathcal{M}\left([Q; \rho]\right)$ and the labeled answer $A$ in previous studies. Then the candidate prompt along with the evaluation score is added to the trajectory for the next iteration. 

The optimization process terminates when the LLM is unable to propose new prompts with better evaluation scores, or a maximum number of optimization steps has reached. 




\subsection{Self-consistency}
\label{sec:definition of SC}

Here, we adopt the definition of self-consistency proposed by \citet{wang2022self}. We sample $n$ responses ($r_1, \cdots, r_n$) from the LLM using the same prompt. The final answer is determined by a voting mechanism, where the most frequent response $a$ is selected as the answer. Self-consistency is the frequency of $a$ in all $n$ responses, which can be formulated as:
\begin{equation}\label{eq:sc}
    \text{SC}=\frac{\sum_{i=1}^n\mathbbm{1}_{a=r_i}}{n}.
\end{equation}

\begin{table*}[!htb]
  \centering
  \small
  \renewcommand\tabcolsep{4pt}
  \begin{tabular}{lccccccccc}
    \toprule
    & \textbf{AddSub} & \textbf{AQuA} & \textbf{Big-Bench Date} & \textbf{GSM8K} & \textbf{MultiArith}  & \textbf{SVAMP} & \textbf{StrategyQA} & \textbf{MATH}\\ 
    \midrule
    \textbf{Correct Answers (\%)} & 96.0 & 79.0 & 83.4 & 82.1 & 97.5 & 90.1 & 95.4 & 70.2 \\
    \textbf{Incorrect Answers (\%)} & 73.4 & 67.1 & 67.8 & 49.3 & 54.2 & 57.5 & 90.6 & 35.9\\
    \textbf{Overall Answers (\%)} & 92.8 & 71.8 & 79.2 & 73.8 & 96.6 & 84.9 & 91.9 & 44.6\\
    \bottomrule
  \end{tabular}
  \caption{The average self-consistency (SC) of correct, incorrect, and overall answers generated by the LLM that prompted with ``Let's think step by step.'' on multiple datasets.}
  \label{tab:self-consistency_different_answers}
  \vspace{2mm}
\end{table*}

\begin{figure*}[!ht]
    \begin{tikzpicture}
        \tikzstyle{mybox} = [draw=white, fill=white, text width=1.5cm, align=center, thin,
                rectangle, rounded corners, fill=GreyBackground]
        \tiny
        \node[mybox](question){Testing\\Question};
        
        \node[right of=question, xshift=1.75cm, mybox](prompt3){\textcolor{red}{Prompt 3}};
        \node[right of=prompt3, xshift=1.75cm, mybox, fill=GreyBackground](answer32){$33, 33$};
        \node[mybox, fill=GreyBackground] at(5.5cm,0.5cm)(answer31){$31$};
        \node[below of=answer31, yshift=-0.05cm, mybox, fill=RedBackground](answer33){$36, \cdots, 36$};
        \node[right of=answer32, xshift=1.75cm, mybox, fill=BlueBackground](self-consistency3){\textcolor{red}{SC:\\ $70.0$}};
        
        \node[above of=prompt3, yshift=0.55cm, mybox](prompt2){Prompt 2};
        \node[above of=answer32, yshift=0cm, mybox, fill=GreyBackground](answer23){$36, 36$};
        \node[above of=answer31, yshift=0.05cm, mybox, fill=GreenBackground](answer22){$31, \cdots, 31$};
        \node[above of=answer23, yshift=0.05cm, mybox, fill=GreyBackground](answer21){$19$};
        \node[right of=answer22, xshift=1.75cm, mybox, fill=BlueBackground](self-consistency2){SC:\\ $70.0$};
        
        \node[above of=prompt3, yshift=1.55cm, mybox](prompt1){Prompt 1};
        \node[above of=answer22, yshift=0cm, mybox, fill=GreenBackground](answer11){$31, \cdots, 31$};
        \node[right of=answer11, xshift=1.75cm, mybox, fill=BlueBackground](self-consistency1){SC:\\ $100.0$};
        
        \node[below of=prompt3, yshift=-0.775cm, mybox](prompt4){Prompt 4};
        \node[below of=answer32, yshift=-0.05cm, mybox, fill=GreyBackground](answer41){$19$};
        \node[below of=answer33, yshift=-0cm,mybox, fill=GreyBackground](answer42){$31, 31, 31$};
        \node[below of=answer41, yshift=-0.05cm, mybox, fill=GreyBackground](answer43){$33, 33$};
        \node[below of=answer42, yshift=-0.1cm, mybox, fill=RedBackground](answer44){$36, \cdots, 36$};
        \node[below of=self-consistency3, yshift=-0.775cm, mybox, fill=BlueBackground](self-consistency4){SC:\\ $40.0$};

        \node[below of=prompt4, yshift=-1.425cm, mybox](prompt5){Prompt 5};
        \node[below of=answer43, yshift=-0.1cm, mybox, fill=GreyBackground](answer51){$8,8$};
        \node[below of=answer44, yshift=-0.05cm,mybox, fill=GreyBackground](answer52){$11$};
        \node[below of=answer51, yshift=-0cm, mybox, fill=GreyBackground](answer53){$19, 19$};
        \node[below of=answer52, yshift=-0.05cm, mybox, fill=GreyBackground](answer54){$31, 31$};
        \node[below of=answer53, yshift=-0.1cm, mybox, fill=RedBackground](answer55){$36,36, 36$};
        \node[below of=self-consistency4, yshift=-1.425cm, mybox, fill=BlueBackground](self-consistency5){SC:\\ $30.0$};

        \node[right of=answer21, xshift=4.75cm, mybox,text width=2.0cm, fill=PurpleBackground](mutual-consistency refinement1){\textcolor{black}{Refinement of\\ Answer} \textcolor{Green}{\textbf{``31''}} };

        \node[right of=self-consistency4, xshift=2.0cm, mybox, text width=2.0cm, fill=PurpleBackground](mutual-consistency refinement2){\textcolor{black}{Refinement of\\ Answer} \textcolor{Red}{\textbf{``36''}}};

        \node[right of=self-consistency1,xshift=5.0cm, mybox,fill=YellowBackground](score1){Score: $87.9$};
        \node[right of=self-consistency2,xshift=5.0cm, mybox,fill=YellowBackground](score2){Score: $81.8$};
        \node[right of=self-consistency3,xshift=5.0cm, mybox,fill=YellowBackground](score3){Score: $50.0$};
        \node[right of=self-consistency4,xshift=5.0cm, mybox,fill=YellowBackground](score4){Score: $45.7$};
        \node[right of=self-consistency5,xshift=5.0cm, mybox,fill=YellowBackground](score5){Score: $44.2$};

        \node[above of=self-consistency1, yshift=0.05cm, mybox, fill=white, text width=3cm](self-consistency evaluation){\small \textcolor{Blue}{\textbf{SC\\ Evaluation (\%)}}};

        \node[right of=self-consistency evaluation, xshift=0.7cm, mybox,fill=white, draw=white, text width=0.3cm]{\small +};

        \node[above of=mutual-consistency refinement1, yshift=0.55cm, mybox, fill=white, text width=2.0cm](mutual-consistency refinement){\small \textcolor{Purple}{\small\textbf{MC \\ Refinement}}};

        \node[right of=self-consistency evaluation, xshift=3.6cm, mybox,fill=white, draw=white, text width=0.3cm]{\small =};

        \node[above of=score1, yshift=0.05cm, mybox, fill=white, text width=2.0cm](GLaPE){\small \textcolor{Yellow}{\textbf{GLaPE\\Metric}}};

        \draw[-stealth, Grey, thick] ([xshift=-0.1cm, yshift=0.4cm]question.east) -- (prompt1.west);
        \draw[-stealth, Grey, thick] ([yshift=0.2cm]question.east) -- (prompt2.west);
        \draw[-stealth, Grey, thick] (question.east) -- (prompt3.west);
        \draw[-stealth, Grey, thick] ([yshift=-0.2cm]question.east) -- (prompt4.west);
        \draw[-stealth, Grey, thick] ([xshift=-0.1cm, yshift=-0.4cm]question.east) -- (prompt5.west);

        \draw[-stealth, Grey, thick] (prompt1.east) -- (answer11.west);
        \draw[-stealth, Grey, thick] ([yshift=0.2cm]prompt2.east) -- (answer21.west);
        \draw[-stealth, Grey, thick] (prompt2.east) -- (answer22.west);
        \draw[-stealth, Grey, thick] ([yshift=-0.2cm]prompt2.east) -- (answer23.west);
        \draw[-stealth, Grey, thick] ([yshift=0.2cm]prompt3.east) -- (answer31.west);
        \draw[-stealth, Grey, thick] (prompt3.east) -- (answer32.west);
        \draw[-stealth, Grey, thick] ([yshift=-0.2cm]prompt3.east) -- (answer33.west);
        \draw[-stealth, Grey, Grey, thick] ([yshift=0.3cm]prompt4.east) -- (answer41.west);
        \draw[-stealth, Grey, thick] ([yshift=0.1cm]prompt4.east) -- (answer42.west);
        \draw[-stealth, Grey, thick] ([yshift=-0.1cm]prompt4.east) -- (answer43.west);
        \draw[-stealth, Grey, thick] ([yshift=-0.3cm]prompt4.east) -- (answer44.west);
        \draw[-stealth, Grey, thick] ([yshift=0.4cm,xshift=-0.1cm]prompt5.east) -- (answer51.west);
        \draw[-stealth, Grey, thick] ([yshift=0.2cm]prompt5.east) -- (answer52.west);
        \draw[-stealth, Grey, thick] (prompt5.east) -- (answer53.west);
        \draw[-stealth, Grey, thick] ([yshift=-0.2cm]prompt5.east) -- (answer54.west);
        \draw[-stealth, Grey, thick] ([yshift=-0.4cm,xshift=-0.1cm]prompt5.east) -- (answer55.west);

        \draw[-stealth, Green, very thick] (answer11.east) -- (self-consistency1.west);
        \draw[-stealth, thick, Grey] (answer21.east) -- ([yshift=0.2cm]self-consistency2.west);
        \draw[-stealth, Green, very thick] (answer22.east) -- (self-consistency2.west);
        \draw[-stealth, thick, Grey] (answer23.east) -- ([yshift=-0.2cm]self-consistency2.west);
        \draw[-stealth, thick, Grey] (answer31.east) -- ([yshift=0.2cm]self-consistency3.west);
        \draw[-stealth, thick, Grey] (answer32.east) -- (self-consistency3.west);
        \draw[-stealth, Red, very thick] (answer33.east) -- ([yshift=-0.2cm]self-consistency3.west);
        \draw[-stealth, thick, Grey] (answer41.east) -- ([yshift=0.3cm]self-consistency4.west);
        \draw[-stealth, thick, Grey] (answer42.east) -- ([yshift=0.1cm]self-consistency4.west);
        \draw[-stealth, thick, Grey] (answer43.east) -- ([yshift=-0.1cm]self-consistency4.west);
        \draw[-stealth, Red, very thick] (answer44.east) -- ([yshift=-0.3cm]self-consistency4.west);
        \draw[-stealth, thick, Grey] (answer51.east) -- ([xshift=0.1cm, yshift=0.4cm]self-consistency5.west);
        \draw[-stealth, thick, Grey] (answer52.east) -- ([yshift=0.2cm]self-consistency5.west);
        \draw[-stealth, thick, Grey] (answer53.east) -- (self-consistency5.west);
        \draw[-stealth, thick, Grey] (answer54.east) -- ([yshift=-0.2cm]self-consistency5.west);
        \draw[-stealth, Red, very thick] (answer55.east) -- ([xshift=0.1cm, yshift=-0.4cm]self-consistency5.west);
        
        \draw[-stealth, thick, Grey] (self-consistency1.east) -- ([yshift=0.1cm]mutual-consistency refinement1.west);
        \draw[-stealth, thick, Grey] (self-consistency2.east) --([yshift=-0.1cm]mutual-consistency refinement1.west);
        \draw[-stealth, thick, Grey] (self-consistency3.east) -- ([yshift=0.2cm]mutual-consistency refinement2.west);
        \draw[-stealth, thick, Grey] (self-consistency4.east) -- (mutual-consistency refinement2.west);
        \draw[-stealth, thick, Grey] (self-consistency5.east) -- ([yshift=-0.2cm]mutual-consistency refinement2.west);        

        \draw[-stealth, thick, Green] ([yshift=0.1cm]mutual-consistency refinement1.east) -- (score1.west);
        \draw[-stealth, thick, Green] ([yshift=-0.1cm]mutual-consistency refinement1.east) -- (score2.west);
        \draw[-stealth, thick, Red] ([yshift=0.2cm]mutual-consistency refinement2.east) -- (score3.west);
        \draw[-stealth, thick, Red] (mutual-consistency refinement2.east) -- (score4.west);
        \draw[-stealth, thick, Red] ([yshift=-0.2cm]mutual-consistency refinement2.east) -- (score5.west);
        
        \draw[-stealth,thick, dashed, red] ([yshift=0.1cm]self-consistency3.east)--([yshift=0.1cm]score3.west);

        \node[above of=mutual-consistency refinement2, yshift=1.15cm, mybox,fill=white, draw=white]{\textcolor{red}{-20.0}};
    \end{tikzpicture}
    \vspace{2mm}
    \caption{The schematic representation of our GLaPE method integrating self-consistency (SC) evaluation and mutual-consistency (MC) refinement. This sketch illustrates how our method assesses the prompts in Figure~\ref{fig:sketch of LLMasOPT}; computation details are provided in Appendix~\ref{sec:computation of GLaPE}. Notably, we observed that prompt3, as indicated by the \textcolor{red}{red} marker, produces an incorrect answer with high self-consistency (70\%). Through the mutual-consistency refinement, our GLaPE score experiences a decrease of $20.0$, rendering it more discernible when compared to prompt1 and prompt2. The texts are favorably read in colors of background. \textcolor{Blue}{Blue}: self-consistency, \textcolor{Purple}{Purple}: mutual-consistency refinement, \textcolor{Green}{Green}: answer ``31'' (gold label), \textcolor{Red}{Orange}: answer ``36'', \textcolor{Yellow}{Yellow}: GLaPE metric.}
    \label{fig:sketch for our GLaPE}
    \vspace{2mm}
\end{figure*}
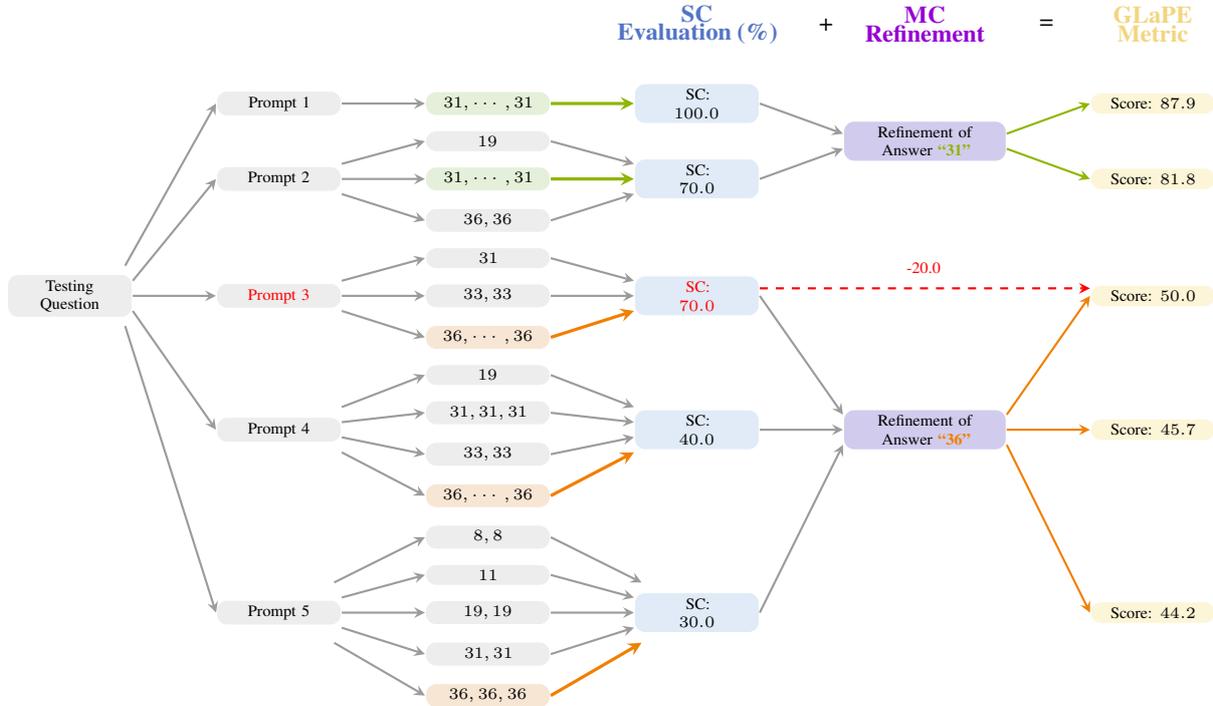

\section{Investigating Gold Label-agnostic Prompt Evaluation}\label{sec:challenge}

According to Section \ref{sec:task_def}, the evaluation function $f$ in existing studies measures the goodness of the prompt candidate $\rho$ by maximizing the task accuracy. However, in real-world tasks, obtaining gold labels poses a considerable challenge, limiting the generalization of existing prompt optimization methods. Furthermore, we ultimately expect LLMs to solve problems for which answers are not already known. Therefore, when optimizing prompts to enhance performance, gold labels are not readily available. Thus, it is imperative to find a gold label-agnostic prompt evaluation method.

In this section, we will investigate the challenge of gold label-agnostic prompt evaluation and study how to design an effective approach to overcome the challenge.




\subsection{SC Fails Due to Overestimating Prompts}\label{sec:solely self-consistency-based prompt optimization}

For a gold label-agnostic prompt evaluation method, it is essential to rely exclusively on the responses and identify patterns within them. 
Building on the findings of \citet{wang2022self}, which demonstrate that selecting the most frequently response enhances accuracy, we aim to investigate whether SC correlates with accuracy.

To this end, we experiment by utilizing the prompt ``Let's think step by step.'' proposed by \citet{kojima2022large}. We calculated the average self-consistency of correct, incorrect, and overall answers and presented the results in Table~\ref{tab:self-consistency_different_answers}. We observe a significant superiority in the average self-consistency of correct answers compared to incorrect ones. A more specific example is shown in Figure \ref{fig:sketch for our GLaPE}. We see that the average SC of correct answers (answer ``31'') significantly surpasses that of incorrect ones. This observation indicates that the self-consistency of responses may reflect accuracy. Thus, it is possible to evaluate prompts based on the SC of the responses and incorporate this method in prompt optimization.

\begin{figure}[!t]
    \centering
    \includegraphics[width=0.4\textwidth]{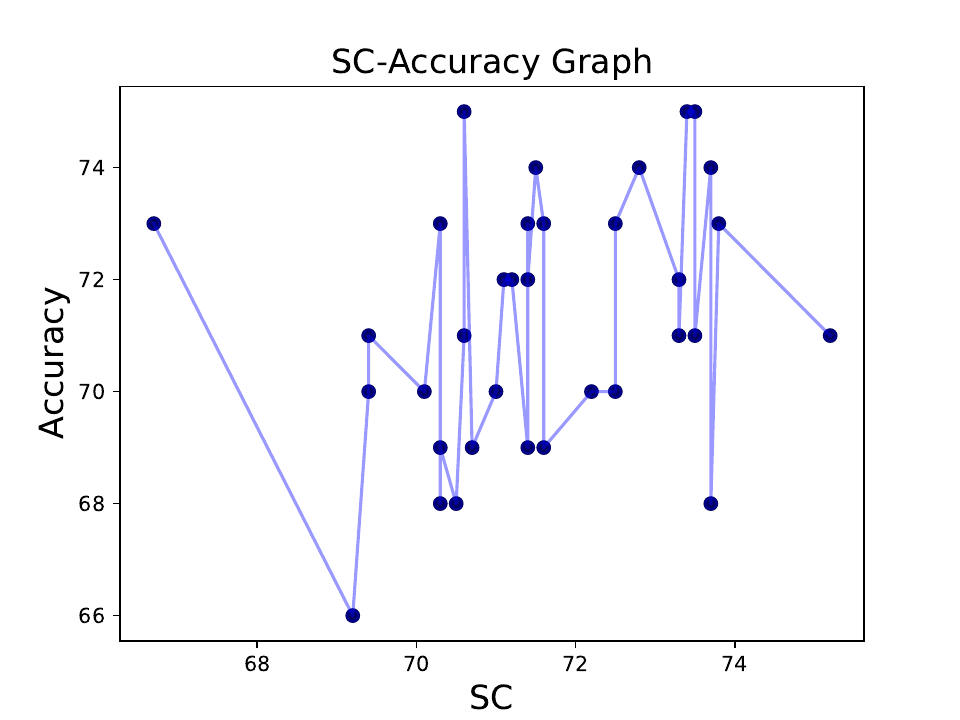}
    \caption{SC-Accuracy Graph for Prompts. Each prompt is represented as a point on the graph, where the x-coordinate signifies self-consistency and the y-coordinate signifies accuracy.}
    \label{fig:self-consistency_based_correlation}
    \vspace{0mm}
\end{figure}

However, we also find that there exists disparity between SC and accuracy when using SC as the sole evaluation metric. This disparity happens to Prompt 3 as shown in Figure \ref{fig:sketch for our GLaPE}. Concretely, Prompt 3 yields an incorrect answer (answer ``36'') but has a high SC of 70.0. 
By taking the GSM8K dataset as the testbed, we computed both the self-consistency and accuracy for a group of prompts. Consequently, we draw each prompt as a point in Figure~\ref{fig:self-consistency_based_correlation}. 
Given the observed fluctuations in the line, it is apparent that self-consistency does not align rigorously with accuracy. Additionally, we find that the Spearman correlation coefficient between SC and accuracy is relatively low, as shown in the first line of Table~\ref{tab:comparison of GLaPE with self-consistency on spearman coefficient}. Therefore, relying on self-consistency alone proves insufficient in offering a comprehensive representation of accuracy in prompt evaluation and optimization.

\begin{table*}[t]
  \centering
  \small
  \renewcommand\tabcolsep{5pt} %
  \begin{tabular}{lcccccccc}
    \toprule
    \textbf{Evaluation Metric} & \textbf{AddSub} & \textbf{AQuA} & \textbf{Big-Bench Date} & \textbf{GSM8K} & \textbf{MultiArith}  & \textbf{SVAMP} & \textbf{StrategyQA} & \textbf{MATH}\\
    \midrule
    GLaPE & 0.44 & 0.04  & 0.88 & 0.49 & 0.88 & 0.69 & 0.18 & 0.67 \\
    SC evaluation & 0.36 & -0.13 & 0.75 & 0.40 & 0.29 & 0.31 & 0.14 & 0.33 \\
    \bottomrule
  \end{tabular}
  \vspace{2mm}
  \caption{Spearman correlation coefficients ($\uparrow$) between accuracy and SC / GLaPE across diverse datasets.}
  \label{tab:comparison of GLaPE with self-consistency on spearman coefficient}
  \vspace{2mm}
\end{table*}


So far, we show that SC alone may not always yield accurate evaluations, since SC does not always align well with accuracy and can overestimate prompts that produce incorrect but consistent answers. Therefore, it deserves a more in-depth investigation to mitigate the side effects of the overestimated prompts by SC. Beyond examining individual prompt responses, we can analyze relationships between different prompts.




\subsection{Mitigating the Challenge with Mutual-consistency (MC) Refinement}

Although the performance of a single prompt is only related to its responses, we leverage other prompts for better evaluation in the absence of a gold label. 

Specifically, we infer the gold label from other prompts and then refine the SC evaluation of the single prompt. Table~\ref{tab:self-consistency_different_answers} shows that correct answers exhibit higher self-consistency (SC), allowing us to predict answer correctness by analyzing the average SC of all prompts producing it.
In Figure~\ref{fig:sketch for our GLaPE}, we can predict that the answer "31" is more likely to be correct, while the answer "36" is not, as the average SC of "31" is 87.5, whereas that of "36" is 46.7. This prediction further aids in refining evaluation of each prompt. 
For an incorrect answer, we should lower the evaluation score of prompts with elevated SC, towards the average.
In Figure~\ref{fig:sketch for our GLaPE}, since the average SC of answer ``36'' is 46.7 while prompt 3 has an elevated SC of 70.0, the evaluation score of prompt 3 should be lowered. This refinement mitigate the SC evaluation of overestimated prompts.

In summary, we predict the correctness of an answer by its average SC and refine each SC towards this average. This aligns the evaluation of prompts producing the same answer.



Based on our pivot study above, we find that combining SC and MC is effective for achieving gold label-agnostic prompt evaluation.

\section{GLaPE}\label{Sec:method}
In light of the discussions in Section \ref{sec:challenge}, we propose GLaPE, a gold label-agnostic prompt evaluation approach. GLaPE is composed of two critical aspects: self-consistency evaluation of a single prompt and mutual-consistency refinement across multiple prompts. The overall procedure is
illustrated in Figure as depicted in Figure~\ref{fig:sketch for our GLaPE}.


For formal description purposes, we assume there are $N$ different prompts and denote the evaluation score for each prompt $\rho_i$ as $f_i$. Among multiple samplings of $\mathcal{M}$ prompted with $\left([Q; \rho_i]\right)$, the answer is $a_i$ and the self-consistency is $c_i$, as defined in Section~\ref{sec:definition of SC}.

\textbf{Self-consistency Evaluation}:
We evaluate prompts based on the self-consistency of their answers by minimizing the loss function:
\begin{equation}
    L_{\text{self}} = \sum_{i=1}^N (f_i - c_i)^2.
\end{equation}

\textbf{Mutual-consistency Refinement}:
Additionally, we propose $L_{\text{refine}}$ as a corrective measure for SC evaluation. It measures and penalizes the mutual inconsistency of evaluation scores ($f_i$) for prompts sharing the same answer:
\begin{equation}
    L_{\text{refine}} = \sum_{1 \leq i < j \leq N} \mathbbm{1}_{a_i=a_j} (f_i - f_j)^2.
\end{equation}

The overall loss function $L_{\text{total}}$ is determined by balancing the loss functions of these two aspects:
\begin{equation}
\label{eq:loss function}
    L_{\text{total}} = \alpha \cdot L_{\text{self}} + (1-\alpha) \cdot L_{\text{refine}},
\end{equation}
where $\alpha$ weights the contribution of self-consistency evaluation and mutual-consistency refinement in the evaluation process. Based on preliminary experiments (detailed in Appendix~\ref{subsec: preliminary experiments}), we set $\alpha = 0.5$.

We obtain the ultimate evaluations $f_1, \cdots, f_N$ by minimizing the loss function $L_{\text{total}}$. We initialize $f_i$ with $c_i$ for simplicity and utilize the default gradient descent method to find the optimal solution with a learning rate of 0.05.

\begin{table*}[!htb]
    \small
    \centering
    \begin{tabular}{lcccc}
        \toprule
        \textbf{Dataset} & \textbf{Method}  & \textbf{Prompt} & \textbf{Accuracy (\%)} \\
        \midrule
        \multirow{4}{*}{Addsub} & Baseline \citep{wang2022self} & \parbox{6cm}{Let's think step by step.} & 85.8 \vspace{3pt} \\
        & OPRO \citep{LLMasOPT} & \parbox{6cm}{Let's meticulously scrutinize every detail.} & 89.4 \vspace{3pt} \\
        & GLaPE-based (Ours)  & \parbox{6cm}{Let's carefully consider each step.} & 87.6 \vspace{3pt} \\
        \midrule
        \multirow{7}{*}{AQuA} & Baseline \citep{wang2022self} & \parbox{6cm}{Let's think step by step.} & 39.4 \vspace{3pt} \\
        & OPRO \citep{LLMasOPT} & \parbox{6cm}{After careful consideration and analysis, the optimal solution is revealed.} & 41.7 \vspace{3pt} \\
        & GLaPE-based (Ours)  & \parbox{6cm}{Through a meticulous analysis of all available data and a strategic approach to problem-solving, a definitive and optimal solution will undoubtedly arise.} & 43.7 \vspace{3pt} \\
        \midrule
        \multirow{6}{*}{Big-Bench Date} & Baseline \citep{wang2022self} & \parbox{6cm}{Let's think step by step.} & 72.4 \vspace{3pt} \\
        & OPRO \citep{LLMasOPT} & \parbox{6cm}{Using a systematic approach and thorough examination, the unequivocal and optimal solution becomes unmistakably evident.} & 72.1 \vspace{3pt} \\
        & GLaPE-based (Ours)  & \parbox{6cm}{Let's analyze this situation thoroughly and explore all possible solutions.} & 71.9 \vspace{3pt} \\
        \midrule
        \multirow{4}{*}{GSM8K} & Baseline \citep{wang2022self} & \parbox{6cm}{Let's think step by step.} & 74.8 \vspace{3pt} \\
        & OPRO \citep{LLMasOPT} & \parbox{6cm}{After careful analysis, the optimal solution becomes clear.} & 76.6 \vspace{3pt} \\
        & GLaPE-based (Ours)  & \parbox{6cm}{After careful analysis, the conclusion is evident.} & 77.7 \vspace{3pt} \\
        \midrule
        \multirow{7}{*}{MultiArith} & Baseline \citep{wang2022self} & \parbox{6cm}{Let's think step by step.} & 98.0 \vspace{3pt} \\
        & OPRO \citep{LLMasOPT} & \parbox{6cm}{Let's approach this problem systematically and strategically, step by step, with logical thinking and methodical planning.} & 99.6 \vspace{3pt} \\
        & GLaPE-based (Ours)  & \parbox{6cm}{Let's approach this problem strategically, methodically, and innovatively, exploring groundbreaking solutions.} & 99.3 \vspace{3pt} \\
        \midrule
        \multirow{7}{*}{SVAMP} & Baseline \citep{wang2022self} & \parbox{6cm}{Let's think step by step.} & 83.9 \vspace{3pt} \\
        & OPRO \citep{LLMasOPT} & \parbox{6cm}{Let's approach this problem with an innovative and revolutionary mindset, breaking the barriers of conventional thinking and achieving unprecedented results.} & 88.9 \vspace{3pt} \\
        & GLaPE-based (Ours)  & \parbox{6cm}{Let's approach this problem with an innovative, revolutionary, and groundbreaking solution.} & 88.7 \vspace{3pt} \\
        \midrule
        \multirow{5}{*}{StrategyQA} & Baseline \citep{wang2022self} & \parbox{6cm}{Let's think step by step.} & 66.1 \vspace{3pt} \\
        & OPRO \citep{LLMasOPT} & \parbox{6cm}{Let's tackle this problem with groundbreaking approaches and unparalleled creativity.} & 69.4 \vspace{3pt} \\
        & GLaPE-based (Ours)  & \parbox{6cm}{Let's explore all the possibilities.} & 70.2 \vspace{3pt} \\
        \midrule
        \multirow{6}{*}{MATH} & Baseline \citep{wang2022self} & \parbox{6cm}{Let's think step by step.} & 21.4 \vspace{3pt} \\
        & OPRO \citep{LLMasOPT} & \parbox{6cm}{Analyzing the data thoroughly can lead to valuable insights.} & 26.4 \vspace{3pt} \\
        & GLaPE-based (Ours)  & \parbox{6cm}{Let's approach this problem with an innovative, revolutionary, and groundbreaking solution.} & 25.9 \vspace{3pt} \\
        \bottomrule
    \end{tabular}
    \vspace{0mm}
    \caption{Optimization results (optimal prompt and corresponding accuracy) of our GLaPE-based prompt optimization method and OPRO \citep{LLMasOPT} across various datasets. Notably, Our optimal prompt is determined by selecting the prompt with the highest GLaPE score.}
    \label{tab:our optimization results on other datasets}
    \vspace{-2mm}
\end{table*}

\section{Experiment}
\subsection{Experiment Setup}\label{Sec:setup}

\textbf{Datasets.} Our experiments were conducted on 8 benchmark datasets to evaluate the performance of our gold label-agnostic prompt evaluation and optimization method. We selected five datasets specifically focused on arithmetic reasoning: AddSub~\citep{hosseini-etal-2014-learning}, AQuA~\citep{ling-etal-2017-program}, GSM8K~\citep{cobbe2021gsm8k}, MultiArith~\citep{roy-roth-2015-solving}, and SVAMP~\citep{patel-etal-2021-nlp}. Additionally, we included the MATH dataset~\citep{hendrycksmath2021}, which is extremely challenging and comprehensive, to test our method's efficacy on particularly difficult benchmarks. Furthermore, we expanded our evaluation to commonsense reasoning benchmarks, such as Big-Bench Date~\citep{srivastava2023beyond} and StrategyQA~\citep{geva2021strategyqa}, to assess the performance of GLaPE in varied contexts.

\textbf{Prompt Optimization.} We implemented the OPRO method proposed by \citet{LLMasOPT} using the prompt shown in Figure 3 of their paper. This technique utilizes an LLM to evaluate existing prompts, generating improved prompts based on the obtained evaluation scores. We chose this approach due to its adaptability; alternative metrics can easily replace evaluation scores in the meta-prompt of optimization. This flexibility facilitates the seamless execution of our gold label-agnostic prompt optimization experiments. Due to time and financial limitations, we conducted both the OPRO and GLaPE-based methods for 16 iterations each, generating 8 prompts per iteration.

\textbf{LLM Backbone.} In both the evaluation and optimization phases, we employed GPT-3.5-turbo-0613, which was the latest version of GPT-3.5-turbo. For prompt evaluation, we empirically set the temperature to $0.7$ and generated $10$ outputs using chain-of-thought prompting \citep{wei2023chainofthought}. For prompt optimization, default hyperparameters and meta-prompt from \citet{LLMasOPT} were applied.

\subsection{Main Results}\label{sec:GLaPE-based prompt optimization}
Table~\ref{tab:our optimization results on other datasets} shows the main results on the 8 benchmark datasets. GLaPE is able to produce effective prompts, achieving performance comparable to those derived from manually annotated gold labels such as OPRO. 
The results suggests that our GLaPE can function as a robust metric, akin to accuracy.
We also compared our method with other recent prompt optimization methods for private LLMs; these results are detailed in Appendix~\ref{subsec:comparison}, providing additional evidence to verify the generality of GLaPE.

\begin{table*}[ht]
    \small
    \centering
    \renewcommand\tabcolsep{16pt} %
    \begin{tabular}{lcc}
        \toprule
        \textbf{Evaluation Metric} & \textbf{Optimal prompt} & \textbf{Accuracy (\%)}   \\
        \midrule
        GLaPE & \parbox{6cm}{After careful analysis, the conclusion is evident.} & 77.7 
         \\
        \addlinespace[2pt]
        SC evaluation & \parbox{6cm}{Let's break it down step by step.} & 75.1 \\
        \bottomrule
    \end{tabular}
    \caption{Comparison of prompt optimization based on self-consistency and our GLaPE.}
    \label{tab:comparison of prompt optimization based on self-consistency and our GLaPE}
    \vspace{1mm}
\end{table*}

\begin{figure*}[!ht]
    \centering
    \begin{subfigure}{0.45\textwidth}
    \centering
        \includegraphics[width=\linewidth]{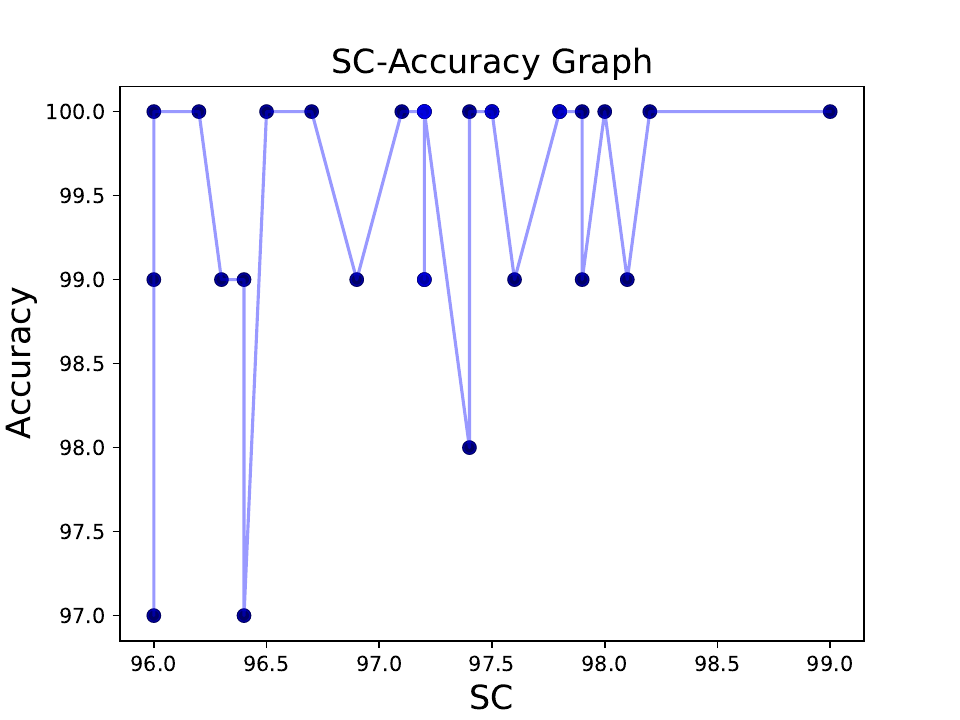}
        \caption{SC-Accuracy Graph.}
        \label{subfig:self-consistency-accuracy graph on multiarith}
    \end{subfigure}
    \hspace{0.05\textwidth}
    \begin{subfigure}{0.45\textwidth}
    \centering
        \includegraphics[width=\linewidth]{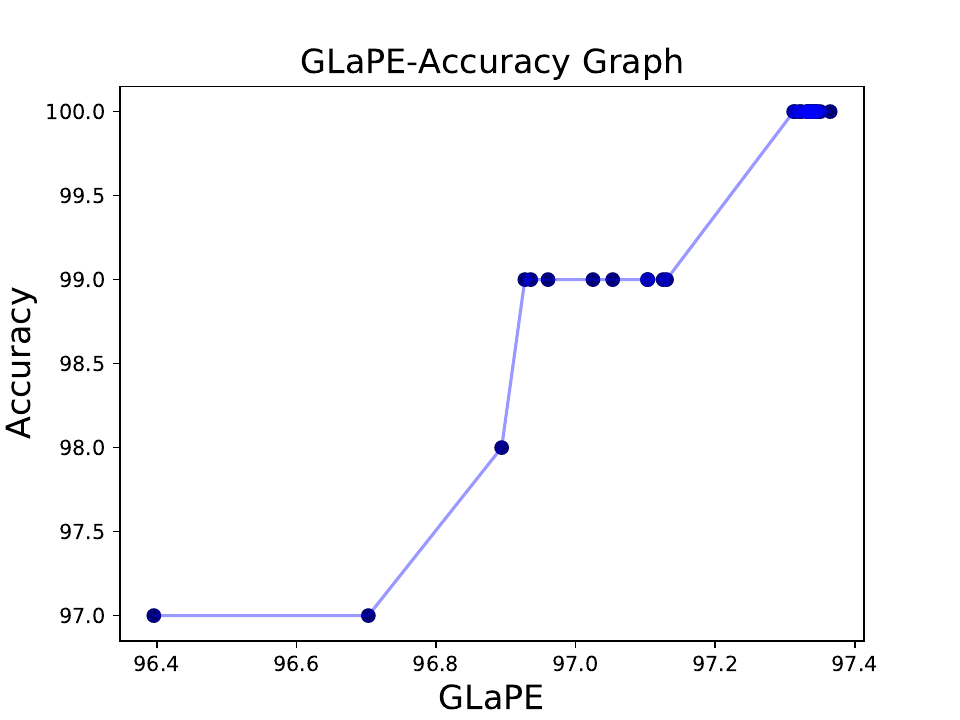}
        \caption{GLaPE-Accuracy Graph.}
        \label{subfig: GLaPE-accuracy graph on multiarith}
    \end{subfigure}
    \caption{SC-Accuracy Graph v.s. GLaPE-Accuracy Graph for Prompts on Multiarith.}
    \label{fig:visualized comparison of GLaPE and self-consistency}
    \vspace{2mm}
\end{figure*}

\subsection{Ablation Study}

\subsubsection{MC Refinement}

In this section, we conduct ablation studies to enhance our understanding of the GLaPE method, with a specific focus on the impact of the mutual-consistency refinement approach.

Initially, on the GSM8K dataset, we compared prompt optimization outcomes using two distinct evaluation methods: self-consistency assessment and GLaPE. As shown in Table~\ref{tab:comparison of prompt optimization based on self-consistency and our GLaPE}, GLaPE-based prompt optimization results in a superior prompt compared to that obtained through confidence assessment. This observation suggests that incorporating mutual-consistency refinement to rectify confidence evaluation enhances the efficacy of prompt optimization.

Furthermore, we incorporated the Spearman correlation coefficient\footnote{\url{https://en.wikipedia.org/wiki/Spearman's_rank_correlation_coefficient}} into our study, wherein a higher coefficient signifies a stronger correlation between variables.
This quantitative assessment was employed to juxtapose GLaPE with the solely SC-based evaluation regarding the correlation with accuracy.
Our analysis concentrated on prompts within the optimization trajectory in the experiment in Section~\ref{sec:GLaPE-based prompt optimization}, to mitigate unnecessary computational costs. 
As shown in Table~\ref{tab:comparison of GLaPE with self-consistency on spearman coefficient}, the Spearman coefficient between GLaPE and accuracy exceeds that of self-consistency across all datasets. 

Additionally, we utilized the visualization method introduced in Section~\ref{sec:solely self-consistency-based prompt optimization} to depict the prompts of the optimization trajectory in a graph (Figure~\ref{fig:visualized comparison of GLaPE and self-consistency}). In Figure~\ref{subfig:self-consistency-accuracy graph on multiarith}, we observe a fluctuating line, whereas in Figure~\ref{subfig: GLaPE-accuracy graph on multiarith}, a consistently increasing line is evident. 
Both of the scrutiny indicate that our mutual-consistency refinement method significantly mitigates the disparity between self-consistency and accuracy.

\subsubsection{Generalizability Across LLMs}

In the experiment of~\ref{sec:GLaPE-based prompt optimization}, we only optimize prompts for GPT-3.5-turbo. To further assess the generalizability of our method on models other than GPT-3.5-turbo, we conducted experiments on three widely used open-source models of different sizes: Mistral-7B, Llama3-8B, and Gemma2-9B. As shown in Table~\ref{tab:results_across_llms}, our GLaPE-based prompt optimization approach consistently discovers prompts that outperform the baselines and are competitive with the OPRO method across all models.

\begin{table}[!ht]
\centering
\small
\begin{tabular}{llc}
\toprule
\textbf{Model} & \textbf{Method} & \textbf{Accuracy (\%)} \\ \midrule
\multirow{3}{*}{\href{https://huggingface.co/mistralai/Mistral-7B-Instruct-v0.2}{Mistral-7B}}     & Baseline & 33.8 \\
               & OPRO                    & 35.9 \\
               & GLaPE (Ours)            & 35.9 \\ \midrule
\multirow{3}{*}{\href{https://huggingface.co/meta-llama/Meta-Llama-3-8B-Instruct}{Llama3-8B}}      & Baseline & 45.4 \\
               & OPRO                    & 48.6 \\
               & GLaPE (Ours)            & 48.9 \\ \midrule
\multirow{3}{*}{\href{https://huggingface.co/google/gemma-2-9b-it}{Gemma2-9B}}      & Baseline & 39.7 \\
               & OPRO                    & 42.4 \\
               & GLaPE (Ours)            & 43.2 \\ \bottomrule
\end{tabular}
\caption{Optimization results of our GLaPE-based prompt optimization method and OPRO \citep{LLMasOPT} across various models on GSM8K dataset.}
\label{tab:results_across_llms}
\vspace{-2mm}
\end{table}

We also investigated the self-consistency (SC) of answers across these models, as presented in Table~\ref{tab:sc_across_llms}. The SC of correct answers consistently exceeds that of incorrect answers for all models, which aligns with the intuition discussed in Section~\ref{sec:solely self-consistency-based prompt optimization}.

\begin{table}[!ht]
\centering
\small
\setlength{\tabcolsep}{3pt} 
\begin{tabular}{lccc}
\toprule
\textbf{Answers} & \textbf{Mistral} & \textbf{Llama3} & \textbf{Gemma2} \\ \midrule
Correct (\%)   & 65.6  & 53.3  & 81.4  \\ \midrule
Incorrect (\%) & 48.3  & 28.8  & 49.6  \\ \midrule
Overall (\%)   & 53.9  & 39.9  & 62.2  \\ \bottomrule
\end{tabular}
\caption{The average self-consistency (SC) of correct, incorrect, and overall answers generated by the multiple LLMs that prompted with ``Let's think step by step.'' on GSM8K datasets.}
\label{tab:sc_across_llms}
\vspace{-3mm}
\end{table}

These results indicate that the our unsupervised GLaPE-based prompt optimization method effectively generalizes across different LLMs, maintaining competitive performance comparing to the supervised OPRO method.

\section{Rethink on Gold Label-agnostic Prompt Optimization}


\label{sec:limitations of our evaluation}

Our amalgamation of self-consistency evaluation and mutual-consistency refinement facilitates the identification of prompts leading to correct answers. However, we also observe a diminished Spearman correlation coefficient between our GLaPE and accuracy on the AQuA dataset and StrategyQA dataset, as depicted in Table~\ref{tab:comparison of GLaPE with self-consistency on spearman coefficient}. Given the suboptimal performance, we shift to reflect on the intrinsic restriction posed by the LLM. As stated in Section~\ref{sec:solely self-consistency-based prompt optimization}, in scenarios where all prompts result in consistent but inaccurate answers, our evaluation may fail to identify the error. 
Without access to external resources, discerning the consistent errors becomes challenging. 
We illustrate some example questions in the Strategy dataset in Figure~\ref{fig:example of consistent error}, where almost all prompts lead to a incorrect answer with extremely high self-consistency during our prompt optimization.
By excluding these problematic questions, we observe a significant improvement in the Spearman correlation coefficient, as detailed in Appendix~\ref{subsec:clean experiment}.

Notably, Our GLaPE evaluation method has been shown effective in challenging datasets, too.
Our GLaPE evaluation method aligns well with accuracy on the rigorous MATH dataset, as shown in Table~\ref{tab:comparison of GLaPE with self-consistency on spearman coefficient}. This alignment can be attributed to the significant disparity in self-consistency between correct and incorrect answers, as depicted in Table~\ref{tab:self-consistency_different_answers}. Conversely, on the StrategyQA dataset, the self-consistency levels of correct and incorrect answers are remarkably similar. This similarity arises from the inherent errors within the LLM itself, which precludes effective distinction between these answers without access to the gold standard labels.

\begin{figure}[!t]
\centering
        \begin{tikzpicture}
            \tikzstyle{mybox} = [draw=white, fill=GreyBackground, thin,
                rectangle, rounded corners, inner sep=5pt, inner ysep=5pt]
            
            \node [mybox] (box1) at (0.0, 0.0){
                \begin{minipage}{0.9\linewidth}                   
                    \small
                    \raggedright

                    \textbf{Example 1:} Was Iggy Pop named after his father?

                    \textbf{Label: } \textcolor{blue}{Yes.}

                    \textbf{LLM Answer \& Explanation:} \textcolor{red}{No.} Iggy Pop, whose real name is James Newell Osterberg Jr., was not named after his father. His father's name was James Newell Osterberg Sr., so they shared the same first name but not the same nickname. Iggy Pop got his stage name from his first band, The Iguanas, and later shortened it to Iggy.

                    \textbf{Example 2:} Is Great Pyramid of Giza the last wonder of its kind?

                    \textbf{Label: } \textcolor{blue}{Yes.}

                    \textbf{LLM Answer \& Explanation:} \textcolor{red}{No.} The Great Pyramid of Giza is one of the Seven Wonders of the Ancient World. It is the only wonder from that list that still exists today. However, there have been other lists of wonders created throughout history. For example, the New7Wonders Foundation conducted a global poll in 2007 to determine the New Seven Wonders of the World. 
                    
                \end{minipage}
            };

        \end{tikzpicture}
        \caption{Some examples in StrategyQA dataset where the LLM consistently provides inaccurate responses.}
        \vspace{2mm}
        \label{fig:example of consistent error}
        \vspace{-0mm}
\end{figure}


\section{Conclusion}

This work presents an innovative gold label-agnostic prompt evaluation methodology that operates in the absence of gold labels. By integrating self-consistency evaluation and mutual-consistency refinement, our evaluation demonstrates a strong correlation with accuracy. Subsequently, we incorporate our metric into prompt optimization, yielding prompts comparable to those optimized based on accuracy metrics across various tasks. 

\section*{Limitations}
First, in Section~\ref{sec:limitations of our evaluation}, we outlined the challenges faced by our GLaPE method in accurately assessing the inherent error of LLM itself. In future research, innovative approaches could be explored to identify the consistent mistakes.
Another limitation in our current evaluation methodology is that we utilize a singular digital score as the assessment, which fails to furnish comprehensive information regarding the prompts. Consequently, future research could augment the granularity of prompt evaluations, incorporating other assessments, like natural language feedback, to address this shortfall.



\bibliography{custom}

\appendix

\section{Appendix}
\label{sec:appendix}

\subsection{Preliminary Experiments}
\label{subsec: preliminary experiments}

In this section, we discuss two crucial hyperparameters used in our experiments. 

The first is the balance weight $\alpha$, which balances SC evaluation and MC refinement as described in Equation~\ref{eq:loss function}. We tested $\alpha$ values of 0.25, 0.5, 0.75, and 1.0, with results detailed in Table~\ref{tab:alpha  test}. An optimal balance was achieved at $\alpha=0.5$, emphasizing the significance of both SC and MC in our evaluation framework. Consequently, we set $\alpha=0.5$ for all experiments.

\begin{table}[!h]
    \small
    \centering
    \renewcommand\tabcolsep{3.5pt}
    \begin{tabular}{lccc}
        \toprule
         \textbf{Weight} $\alpha$  & \textbf{Prompt} & \textbf{Accuracy (\%)} \\
        \midrule
        0.25 & \parbox{3.5cm}{Let's think about this logically.} & 77.2 \vspace{3pt} \\
        0.5 & \parbox{3.5cm}{After careful analysis, the conclusion is evident.} & 77.7 \vspace{3pt} \\
        0.75 & \parbox{3.5cm}{Let's approach this problem with utmost creativity, innovation, and strategic thinking.} & 76.4 \vspace{3pt} \\
        1.0 & \parbox{3.5cm}{Let’s break it down step by step.} & 75.1 \vspace{3pt} \\
        \bottomrule
    \end{tabular}
    \caption{Optimization results on the GSM8K dataset using different values of balance weight $\alpha$ as specified in Equation~\ref{eq:loss function}.}
    \label{tab:alpha test}
    \vspace{-3mm}
\end{table}

The second parameter is the training dataset size. We evaluated various sizes: 10, 20, 50, 100, and 200, as shown in Table~\ref{tab:dataset size test}. Based on these results, we selected a dataset size of 100 to balance accuracy and computational efficiency.

\begin{table}[!h]
    \small
    \centering
    \renewcommand\tabcolsep{3pt}
    \begin{tabular}{lccc}
        \toprule
         \textbf{Dataset Size}  & \textbf{Prompt} & \textbf{Accuracy (\%)} \\
        \midrule
        10 & \parbox{3.5cm}{Let's break it down step by step.} & 75.1 \vspace{3pt} \\
        20 & \parbox{3.5cm}{Let's carefully analyze each aspect of the problem thoroughly and devise the most optimal plan.} & 75.5 \vspace{3pt} \\
        50 & \parbox{3cm}{Let's approach this problem with utmost creativity, innovation, and strategic thinking.} & 76.4 \vspace{3pt} \\
        100 & \parbox{3.5cm}{After careful analysis, the conclusion is evident.} & 77.7 \vspace{3pt} \\
        200 & \parbox{3.5cm}{Let’s break it down step by step.} & 77.9 \vspace{3pt} \\
        \bottomrule
    \end{tabular}
    \caption{Optimization results on the GSM8K dataset using different training dataset sizes.}
    \label{tab:dataset size test}
    \vspace{-3mm}
\end{table}

\begin{table*}[!h]
    \small
    \centering
    \begin{tabular}{lcccc}
        \toprule
        \textbf{Dataset} & \textbf{Method}  & \textbf{Prompt} & \textbf{Accuracy (\%)} \\
        \midrule
        \multirow{11}{*}{GSM8K} & Baseline \citep{wang2022self} & \parbox{6cm}{Let's think step by step.} & 74.8 \vspace{3pt} \\
        & APE~\citep{zhou2022large} & \parbox{6cm}{Let’s work this out in a step by step way to be sure we have the right answer.} & 76.3 \vspace{3pt} \\
        & APO~\citep{pryzant2023automatic} & \parbox{6cm}{Given the scenario, perform necessary calculations and provide a step-by-step explanation to arrive at the correct numerical answer. Consider all information provided.} & 76.5 \vspace{3pt} \\
        & PE2~\citep{ye2023prompt} & \parbox{6cm}{Let’s solve the problem step-by-step and calculate the required total value correctly.} & 77.7 \vspace{3pt} \\
        & GLaPE-based (Ours)  & \parbox{6cm}{After careful analysis, the conclusion is evident.} & 77.7 \vspace{3pt} \\
        \midrule
        \multirow{13}{*}{MultiArith} & Baseline \citep{wang2022self} & \parbox{6cm}{Let's think step by step.} & 98.0 \vspace{3pt} \\
        & APE~\citep{zhou2022large} & \parbox{6cm}{Let’s work this out in a step by step way to be sure we have the right answer.} & 97.8 \vspace{3pt} \\
        & APO~\citep{pryzant2023automatic} & \parbox{6cm}{Given the scenario, perform the necessary calculations step by step to find the final result. Consider all parts of the input and the sequence of events.} & 99.0 \vspace{3pt} \\
        & PE2~\citep{ye2023prompt} & \parbox{6cm}{Let’s solve this problem by considering all the details. Pay attention to each piece of information, remember to add or subtract as needed, and perform the calculations step by step} & 99.6 \vspace{3pt} \\
        & GLaPE-based (Ours)  & \parbox{6cm}{Let's approach this problem strategically, methodically, and innovatively, exploring groundbreaking solutions.} & 99.3 \vspace{3pt} \\
        \bottomrule
    \end{tabular}
    \caption{Optimization results (optimal prompt and corresponding accuracy) of our GLaPE-based prompt optimization method and other popular methods. }
    \label{tab:method comparison}
    \vspace{-3mm}
\end{table*}

\subsection{Computation Detail of Figure~\ref{fig:sketch for our GLaPE}}
\label{sec:computation of GLaPE}

First, we calculate the self-consistency $c_i$ for each prompt according to the definition in Section~\ref{sec:definition of SC}, which are:
\begin{align*}
    &c_1 = 100.0, \quad c_2 = 70.0, \quad c_3 = 70.0, \\
    &c_4 = 40.0, \quad c_5 = 30.0.
\end{align*}

Thus, the loss function $L_{\text{self}}$ is:
\begin{align*}
L_{\text{self}} &= \sum_{i=1}^{5} (f_i - c_i)^2 = (f_1 - 100)^2 + (f_2 - 70)^2 \\
&\quad + (f_3 - 70)^2 + (f_4 - 40)^2 + (f_5 - 30)^2.
\end{align*}

Next, we calculate the loss function of mutual-consistency refinement $L_{\text{refine}}$, which is:
\begin{align*}
L_{\text{refine}} = \sum_{1 \leq i < j \leq 5} \mathbbm{1}_{a_i = a_j} (f_i - f_j)^2,
\end{align*}
since prompts 1 and 2 share the same answer 31, while prompts 3, 4, and 5 share the same answer 36.

Clearly, $f_1$ and $f_2$ are unrelated to $f_3$, $f_4$, and $f_5$ since their answers are different.

The evaluation scores are then computed as follows (ignoring the coefficient $0.5$ for both $L_{\text{self}}$ and $L_{\text{refine}}$):
\begin{align*}
f_1, f_2 &= \arg \min_{f_1, f_2} \big[ (f_1 - 100)^2 + (f_2 - 70)^2 \\
&\quad + (f_2 - 70)^2 + (f_1 - f_2)^2 \big]
\end{align*}
and 
\begin{align*}
f_3, f_4, f_5 &= \arg \min_{f_3, f_4, f_5} \big[ (f_3 - 70)^2 + (f_4 - 40)^2 \\
&\quad + (f_5 - 30)^2 + (f_3 - f_4)^2 + (f_3 - f_5)^2\\
&\quad + (f_4 - f_5)^2 \big].
\end{align*}

Ultimately, the solution is:
\begin{align*}
    &f_1 = 87.9, \quad f_2 = 81.8, \quad f_3 = 50.0, \\
    &f_4 = 45.7, \quad f_5 = 44.2.
\end{align*}

\subsection{Further Comparison of Prompt Optimization Methods}
\label{subsec:comparison}
To emphasize the efficacy of our method, we conducted additional comparisons between our GLaPE method and other recent prompt optimization approaches for private LLMs, including APE~\citep{zhou2022large}, APO~\citep{pryzant2023automatic}, and PE2~\citep{ye2023prompt}. The results are presented in Table~\ref{tab:method comparison}. These comparisons demonstrate that GLaPE is not only competitive but also exceeds the performance of other existing supervised methods in various cases.

\subsection{Spearman Correlation Coefficients on Cleaned Datasets}
\label{subsec:clean experiment}
It is imperative to recognize that our methodology evaluates prompts on individual questions, and the evaluation score of a prompt across the entire dataset is derived from the sum of its evaluation scores on each question. Consequently, inaccuracies in evaluations for questions stated in Section~\ref{sec:limitations of our evaluation} can significantly compromise the effectiveness of the overall dataset evaluation, particularly on challenging datasets. To gauge the impact of challenging questions on our GLaPE, we exclude questions for which no prompt results in a correct answer with a self-consistency level greater than $50\%$ from the dataset. The cleaned dataset was then compared to a control group, consisting of an equally large subset of the original dataset, to mitigate the influence of dataset size bias. On the initial dataset, the control group, and the cleaned dataset, we calculate the Spearman correlation coefficient. 

In Table~\ref{tab:eliminating intrinsic limitation of LLM for evaluation}, the Spearman correlation coefficient on the cleaned dataset demonstrates a considerable improvement compared to that on the original dataset or control group. This improvement underscores the pronounced adverse influence of intricate questions on our evaluation process. 

\begin{table*}[t]
  \centering
  \small
  \renewcommand\tabcolsep{2pt} %
  \begin{tabular}{lcccccccc}
    \toprule
    & \textbf{AddSub} & \textbf{AQuA} & \textbf{Big-Bench Date} & \textbf{GSM8K} & \textbf{MultiArith}  & \textbf{SVAMP} & \textbf{StrategyQA} & \textbf{MATH}\\
    \midrule
    \textbf{Cleaned Dataset} & 0.61(+0.17) & 0.40(+0.36) & 0.94(+0.06) & 0.69(+0.20) & 0.93(+0.05) & 0.81(+0.12) & 0.41(+0.13) & 0.61(+0.14) \vspace{1mm}\\
    \textbf{Control Group} & 0.42(-0.07) & -0.01(-0.05) & 0.86(-0.02) & 0.40(-0.09) & 0.84(-0.04) & 0.61(-0.08) & 0.16(-0.02) & 0.46(-0.01) \vspace{1mm}\\
    \textbf{Original Dataset} & 0.44 & 0.04 & 0.88 & 0.49 & 0.88 & 0.69 & 0.18 & 0.47 \\
    \bottomrule
  \end{tabular}
  \vspace{-1mm}
  \caption{Comparison of Spearman correlation coefficients ($\uparrow$) before and after excluding challenging questions that surpass the intrinsic capabilities of LLM. Evaluation of the control group is conducted by randomly selecting 10 subsets of the original dataset, and the average Spearman correlation coefficient is computed.}
  \label{tab:eliminating intrinsic limitation of LLM for evaluation}
  \vspace{-5mm}
\end{table*}


\end{document}